\documentclass[conference,compsoc]{IEEEtran}

\usepackage{booktabs}
\usepackage[cmex10]{amsmath}
\usepackage{graphicx}
\usepackage{subfig}
\usepackage{tabularx}

\newcommand{\calX}{\mathcal{X}}

\begin{document}
\title{Scalable Exact Parent Sets Identification in Bayesian Networks Learning\\ with Apache Spark}

\author{\IEEEauthorblockN{Subhadeep Karan}
  \IEEEauthorblockA{Department of Computer Science and Engineering\\
    University at Buffalo\\
    Buffalo, NY, USA\\
    Email: skaran@buffalo.edu}
  \and
  \IEEEauthorblockN{Jaroslaw Zola}
  \IEEEauthorblockA{
    Department of Computer Science and Engineering\\
    Department of Biomedical Informatics\\
    University at Buffalo\\
    Buffalo, NY, USA\\
    Email: jzola@buffalo.edu}
}
\maketitle

\begin{abstract}
In Machine Learning, the parent set identification problem is to find a set of random variables that best explain selected variable given the data and some predefined scoring function. This problem is a critical component to structure learning of Bayesian networks and Markov blankets discovery, and thus has many practical applications, ranging from fraud detection to clinical decision support.

In this paper, we introduce a new distributed memory approach to the exact parent sets assignment problem. To achieve scalability, we derive theoretical bounds to constraint the search space when MDL scoring function is used, and we reorganize the underlying dynamic programming such that the computational density is increased and fine-grain synchronization is eliminated. We then design efficient realization of our approach in the Apache Spark platform. Through experimental results, we demonstrate that the method maintains strong scalability on a 500-core standalone Spark cluster, and it can be used to efficiently process data sets with 70 variables, far beyond the reach of the currently available solutions.
\end{abstract}

\IEEEpeerreviewmaketitle

\section{Introduction}

In Machine Learning, the parent set assignment problem is to find a set of random variables that best explain a selected variable given input data and some predefined scoring criterion~\cite{Koivisto2006}. It is a precursor to Bayesian networks structure learning, where it is solved for each variable to produce a list of potential predecessors of that variable in a final network, which reduces the number of structures that have to be considered~\cite{Yuan2011,Scanagatta2015,Karan2016}. It is also closely related to the feature selection problem, since it directly translates into Markov blankets discovery~\cite{Tsamardinos2003}. Because of these connections, the problem has many practical applications spanning clinical decision support systems, risk assessment, strategic planning, fraud detection and many others~\cite{Kong2008,Fenton2010,Mukhanov2008,Friedman2000}. In all these applications, random variables model attributes of interest, their realizations are observed from~data, and a model obtained by solving the parent set assignment provides insights into how different attributes depend on each other, including conditional dependencies.

While the parent set identification is critical to building models like Bayesian networks, it is known to be formally hard for the most commonly used scoring functions. For instance, it is $\mathcal{NP}$-complete for the Normalized Maximum Likelihood (NML) criterion~\cite{Koivisto2006}. Consequently, the current approaches, which we briefly review in Section~\ref{sec:prior}, either depend on heuristics or deliver exact solutions but are limited in how large instances they can solve. In fact, the largest problems solved by exact algorithms do not contain more than 40 variables~\cite{Fan2015}. In contrast, the real-world systems that strongly depend on the high quality Bayesian networks often involve hundreds of variables. The available heuristics that can solve instances of that size, e.g.~\cite{Scanagatta2015}, do not provide any guarantees on the quality of the solutions they find. This significantly impacts their usefulness, since the inherent uncertainty of the model due to the input data and the scoring function, cannot be separated from the deficiencies of the learning algorithm~\cite{Koller2009}. Consequently, there is a gap between the quality and the size of the models that depend on the exact parent set identification and that can be efficiently learned from the data.

Responding to the above challenges, in this paper, we propose a new distributed memory algorithm for the exact parent sets identification problem. Our goal is to push the limit on the scale of instances that can be solved in acceptable time limits on a modestly sized parallel cluster. To this end, we make the following specific contributions: 1) we propose a new strategy to constraint and reorganize dynamic programming computations in the parent set assignment problem such that computational grain is improved and fine-grained synchronization is avoided, 2) we define a simple mechanism that we use to change the mode of computations from BFS to DFS such that the main memory is preserved. To validate our approach, we provide an efficient implementation on the Apache Spark platform~\cite{Zaharia2016}, and demonstrate its strong scalability across different ML test sets. We then show that on a 500-core cluster with 25 nodes, our system can process HEPAR II test data~\cite{Onisko2000} in slightly over 20 hours. With 70 variables and small variability, this data set is one of the most challenging benchmarks for Bayesian networks structure learning, and it has no exact results available to~date.
 

The paper is organized following the common practice. In Section~\ref{sec:problem}, we provide basic definitions and formally state the problem. In Section~\ref{sec:method}, we introduce our proposed method, and we describe its experimental validation in Section~\ref{sec:results}. We summarize related work in Section~\ref{sec:prior}, and conclude the paper in Section~\ref{sec:conclusion}.

\section{Preliminaries}\label{sec:problem}

Consider a set of $n$ random variables $\calX = \{X_1, \ldots, X_n \}$, and suppose that we are given a complete input data table $D = \{ D_1,\ldots,D_n\}$, where $D_i$ is a vector of $m$ observations of $X_i$. Let $s(X_i, Pa(X_i))$ be a scoring function that quantifies how well $X_i$ is explained by a set of variables $Pa(X_i) \subseteq \calX - \{X_i\}$ given the data~$D$. We will call $Pa(X_i)$ a parent set, or simply parents, of~$X_i$. We are assuming that $s$ is given. For example, it could be one of the several available functions, such as information theoretic MDL~\cite{Schwarz1978} and AIC~\cite{Akaike1973} or the BD family implementing Bayesian scoring criteria~\cite{Cooper1992,Heckerman1999}. In this work, we mostly focus on the MDL scoring function, but our results can be generalized to other scoring criteria. Moreover, we do not consider details of how $s$ is computed, except that it has to access the data in $D$, and the cost of computations is not negligible as it grows with the size of $Pa(X_i)$ and the number of observations $m$. The {\it parent set assignment problem} is to find a subset $Pa(X_i) \subseteq \calX - \{X_i\}$ such that $s(X_i, Pa(X_i))$ is minimized.

Let $d(X_i, U), U \subseteq \calX-\{X_i\}$, be the score of selecting {\it optimal parent set} of $X_i$ from among variables in $U$, that is $d(X_i, U) = \displaystyle\min_{Pa(X_i) \subseteq U} s(X_i, Pa(X_i))$. We can efficiently express $d$ via the following recursion:
\begin{equation}\label{eq:d}
d(X_i,U) = \min 
  \begin{cases}
  s(X_i,U), \\
  \displaystyle\min_{X_j \in U} d(X_i,U-\{X_j\}).
  \end{cases}
\end{equation}
To find an optimal parent set assignment of $X_i$ we could solve the recursion in Eq.~(\ref{eq:d}) for $U = \calX - \{X_i\}$ while recording the choice of parents we made in the process. However, in the majority of practical applications, especially in the context of Bayesian networks structure learning, it is necessary to consider a slightly broader version of the problem.

We will say that $U \subseteq \calX - \{X_i\}$ is a {\it maximal parent set} of $X_i$ if $d(X_i,U) = s(X_i,U)$. From Eq.~(\ref{eq:d}) we have that if $U$ is a maximal parent set then no subset of $U$ has score better than $d(X_i, U)$, i.e. $\displaystyle\forall_{U' \subset U} d(X_i,U) < d(X_i,U')$. Hence, by identifying all maximal parent sets of $X_i$ and memoizing their corresponding scores $s$, we can efficiently answer queries about any optimal parent set of $X_i$. Specifically, to answer query $d(X_i,U')$ for any $U'$ it is sufficient to check if $U'$ is one of the maximal parent sets of $X_i$. If it is, then all we have to do is to return the memoized score $s$ of that maximal parent set. Otherwise, $d(X_i,U')$ must be equal to the smallest $s$ among all maximal parent sets of $X_i$ for which $U'$ is a superset.

The above property of maximal parent sets has important practical implications. For example, to compute the score $Q(\calX)$ of an optimal Bayesian network over $\calX$, and thus find the network itself, we have to solve recursion of the form $Q(U) = \displaystyle\min_{X_i \in U}(d(X_i,U - \{X_i\}) + Q(U - \{X_i\}))$. Even with the efficient algorithms such as~\cite{Yuan2011,Karan2016} this requires large and hard to predict number of queries for optimal parent sets, owing to the component $d(X_i,U - \{X_i\})$ in the recursion. Because for a single variable $X_i$ there are $2^{n-1}$ optimal parent sets, memoizing them all is impractical and often infeasible. In contrast, the set of all maximal parent sets is usually many orders of magnitude smaller, and hence using it instead, in the way we explained before, is the desired and viable alternative~\cite{Yuan2011,Karan2016,Scanagatta2015}.

From the computational point of view, identifying maximal parent sets of $X_i$ is the same as selecting its optimal parent set from $\calX - \{X_i\}$. The only difference are extra steps required to test and store subsets that correspond to maximal parent sets. In practical settings, we wish to enumerate all maximal parent sets for all variables in $\calX$, and this is the problem we are considering in this paper.

\section{Proposed Approach}\label{sec:method}

\begin{figure*}
\centering
\subfloat[]{\label{fig:space_dag}\includegraphics[scale=0.33]{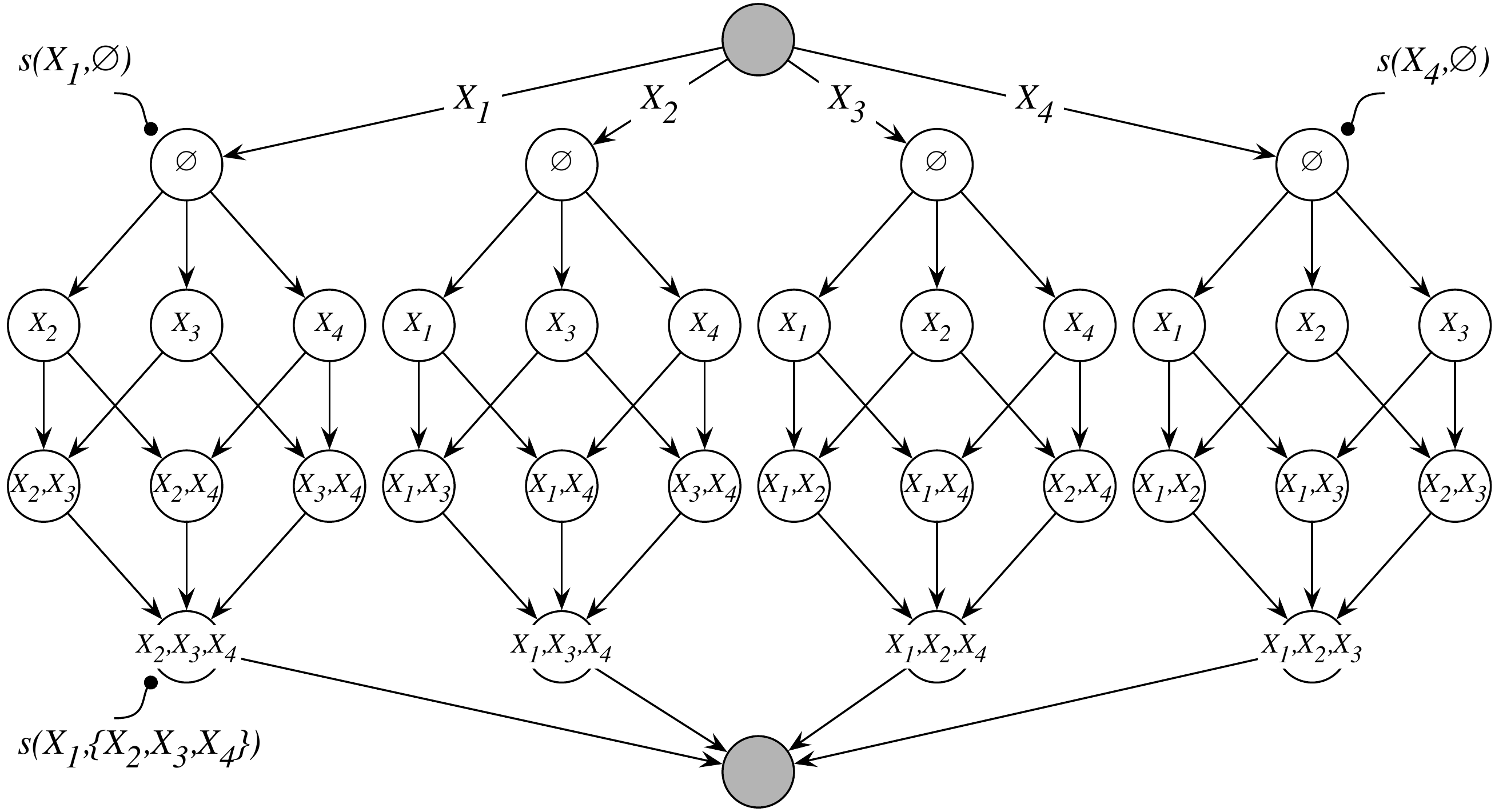}}~~~~\subfloat[]{\label{fig:space_pruned}\includegraphics[scale=0.33]{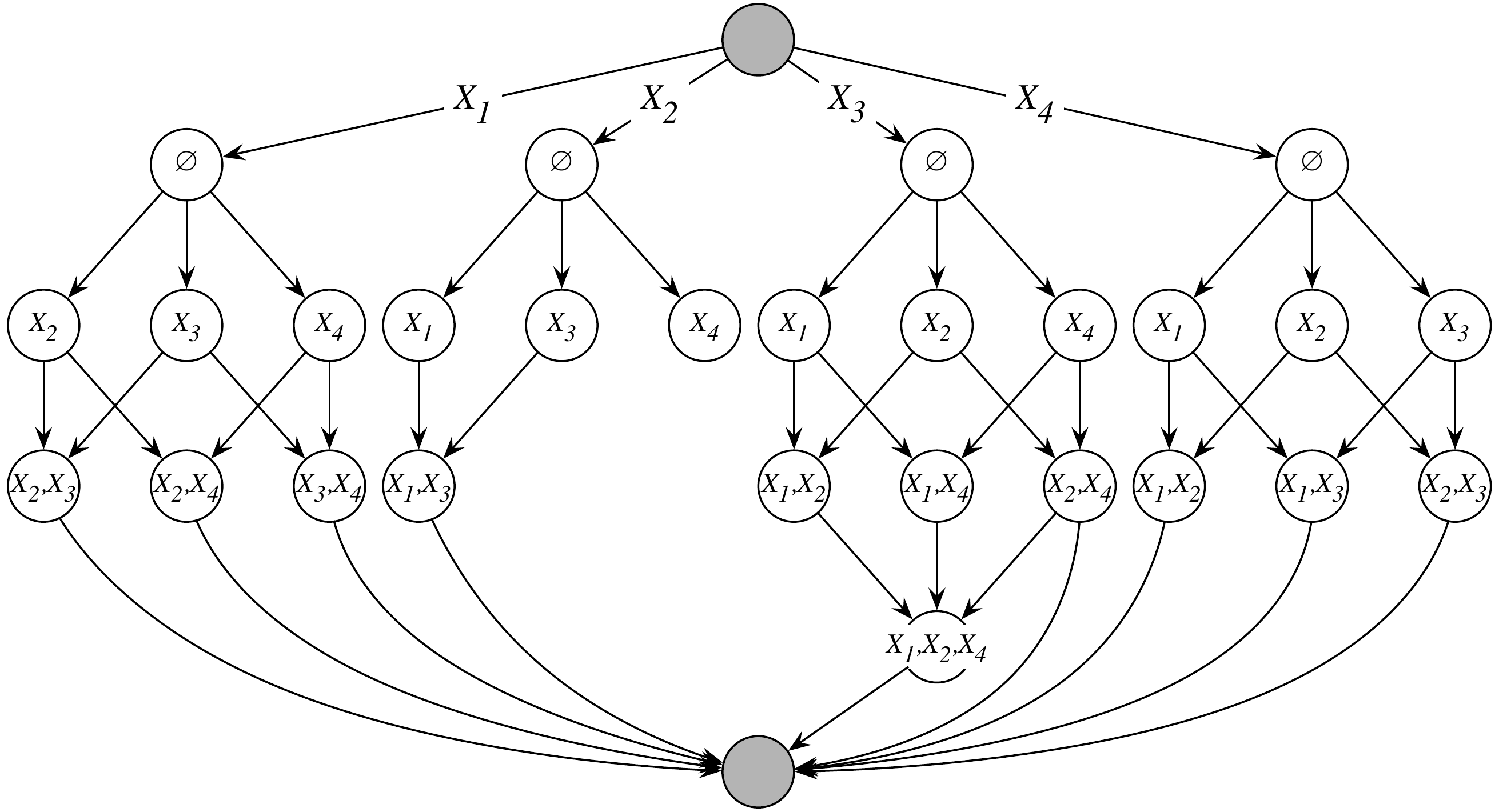}}\\
\subfloat[]{\label{fig:space_folded}\includegraphics[scale=0.33]{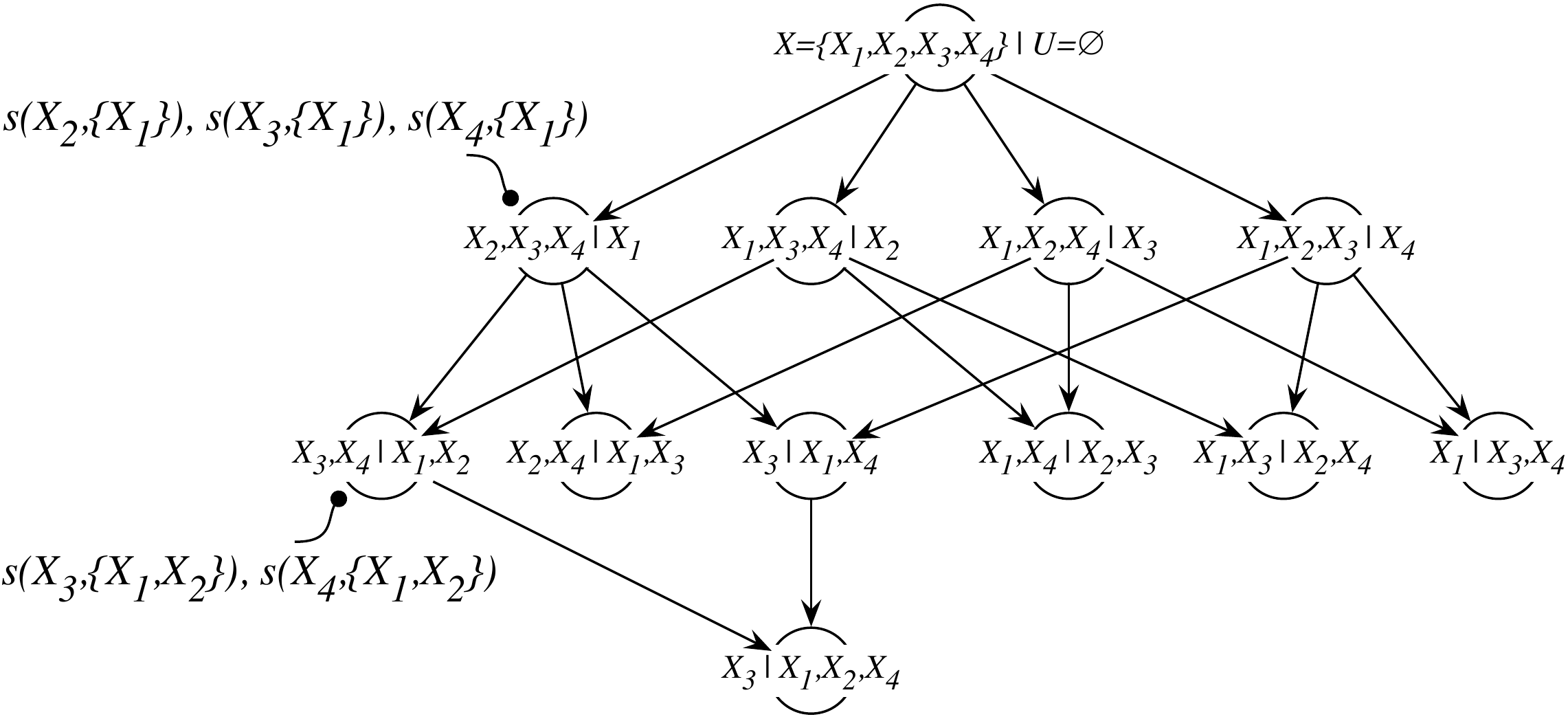}}~~~~~~\subfloat[]{\label{fig:space_reordered}\includegraphics[scale=0.33]{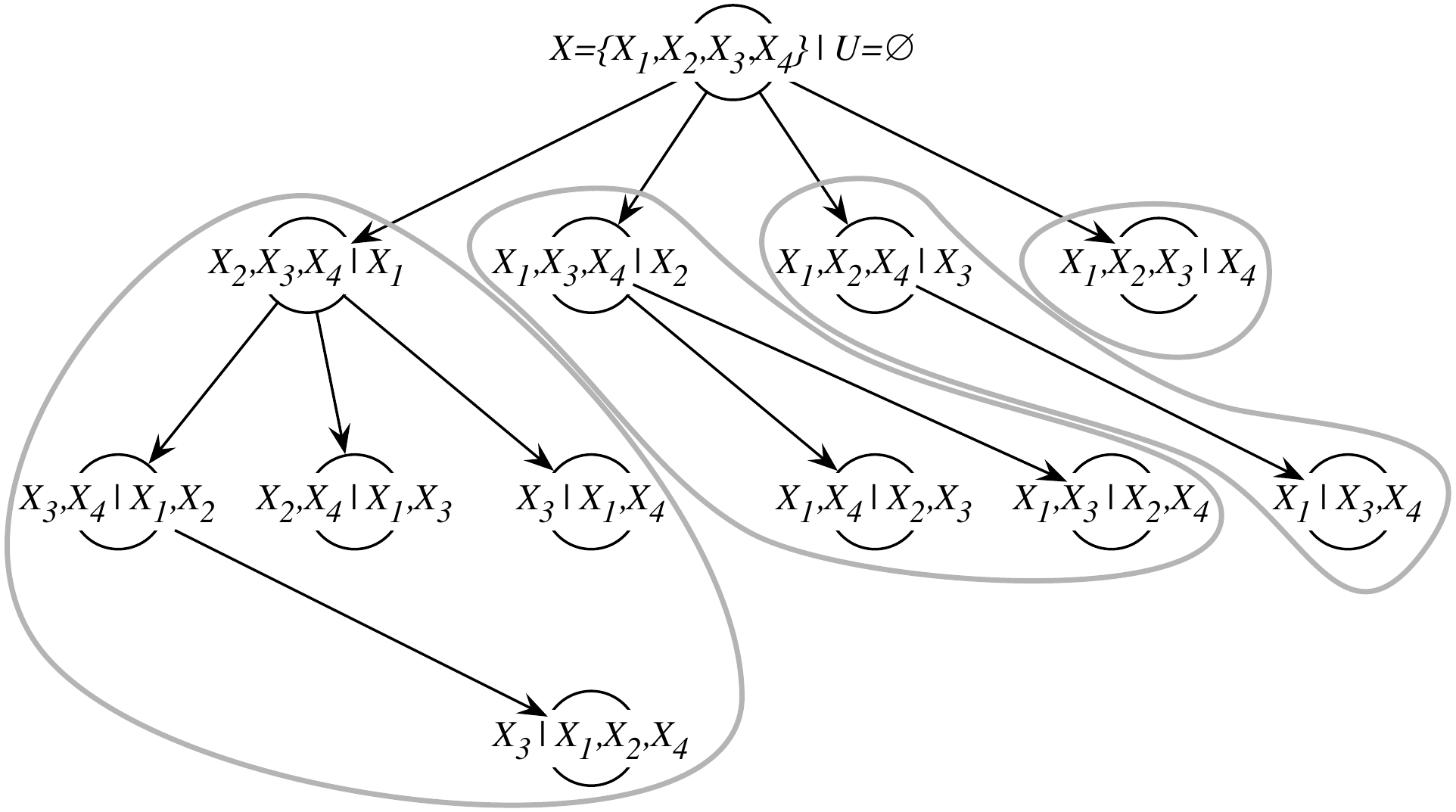}}
\caption{(a) Example of the dynamic programming lattices for $\calX=\{X_1, X_2, X_3, X_4\}$. Processing node $U$ in a lattice for variable $X_i$ requires computing $s(X_i, U)$ and access to $d(X_i, U')$ for each predecessor $U'$ of $U$. (b) Example constrained lattices, and (c) their corresponding ``folded'' representation. A node $U$ in the compacted lattice requires that $s$ is evaluated for several variables that share candidate parents $U$. This improves efficiency of computing $s$, decreases memory requirements and increases computational density. (d) Precedence constraints after eliminating fine-grain synchronization within every layer of the lattice in~(c). Nodes processed by the same task if the DFS mode is initiated at layer $l=1$ are~outlined. Note that following how input variables are ordered, the nodes in the larger tasks are more likely to be pruned.}\label{fig:space}
\end{figure*}

Given a set of variables $\calX$, database of observations~$D$, and a scoring function $s$, our goal is to enumerate all maximal parent sets for all $X_i \in \calX$. If we consider a single variable $X_i$, then we can directly apply recursion in Eq.~(\ref{eq:d}) and starting from the empty set we can consider parent sets of growing size. This process can be though of as a top-down traversal of the dynamic programming lattice with $n$ levels formed by the partial order ``set inclusion'' on the power set of $\calX - \{X_i\}$ (see Figure~\ref{fig:space_dag}). At the level $l=0$ of the lattice we have empty set. Two nodes in the lattice, $U'$ and $U$, are connected only if $U' \subset U$ and $|U| = |U'| + 1$. Here we use $U$ to denote both a subset of $\calX - \{X_i\}$ and the corresponding node in the lattice. A node $U$ represents a parent set of $X_i$. When it is discovered, we compute $s(X_{i}, U)$, compare it with scores $d$ passed by its predecessors to obtain $d(X_i, U)$, and if $U$ is a maximal parent set we store or report a tuple $(X_i, U, s(X_i, U))$.

While the above strategy is clearly guaranteed to enumerate all maximal parent sets, it is both computationally and memory challenging. The computational complexity is due to the $\Theta(2^n)$ invocations of $s$, and memory complexity is driven by how the dynamic programming lattice is traversed. For example, one way is to assume BFS traversal induced by the precedence constraints in the lattice. In such case, maximal parent sets are enumerated layer-by-layer with a synchronization point between any two consecutive layers. This strategy requires that both layers are stored in the memory, which implies $O\left(\binom{n}{\frac{n}{2}}\right)$ space complexity, irrespective of which parallel BFS realization we assume~\cite{Leiserson2010}. Another approach is to use some variant of DFS. With DFS we can benefit from techniques like hypercube pipelining, similar to~\cite{Nikolova2013}, but this strategy requires that we store partial results and update them each time a node is discovered before its all predecessors are processed. As a result, the space complexity is $O(2^{n})$ and we have to maintain potentially irregular memory updates to detect new maximal parent sets. Finally, while all variables in $\calX$ can be processed independently, the resulting embarrassing parallelism is highly limited. This is because the computational cost for a single variable is exponential in~$n$, which effectively constraints the total number of variables we may hope to process. For example, if we assume $n=48$ then the estimated memory requirements to process a single lattice, with $O(2^n)$ nodes and 16~B per node, is 4~PB with a modest 48-way parallelism.

\subsection{Constraining the Search Space}

To achieve a scalable strategy, we start from constraining the search space. This is necessary since the exponential cost of considering all optimal parent sets is prohibitive for realistic problem instances, irrespective of how efficient is our parallel exploration algorithm.

For every variable $X_i$ it is reasonable to expect that its optimal parent set will not contain all other variables. In other words, there is a limit on the depth to which we should be exploring the dynamic programming lattice of~$X_i$. To maintain exactness guarantees, we have to ensure that the bound on the depth of exploration is no smaller than the unknown size of the optimal parent set. Here we provide such a bound for the information-theoretic MDL scoring function. We note that similar bounds can be derived for other functions, and in fact a significant work in this direction has been done, for example~in~\cite{deCampos2011}.

The MDL score is defined as:
\begin{equation}\label{eq:mdl}
s(X_i, U) = m\cdot\mathcal{H}(X_i|U) + \mathcal{NC}(X_i, U),
\end{equation}
where
\begin{equation}\label{eq:nc}
\mathcal{NC}(X_i, U) = \frac{1}{2}\cdot\log_2(m) \cdot q_i \cdot (r_i - 1)
\end{equation}
is a network complexity term, $\mathcal{H}(X_i|U)$ is the estimated conditional entropy of $X_i$ given $U$, and $r_i > 1$ is the number of states of $X_i$, and $q_i = \displaystyle\prod_{j,X_j \in U} r_j$ is the number of states that variables in $U$ can assume ($q_i = 1$ if $U=\emptyset$). The parameters $r_i$ and $q_i$ as well as the conditional entropy are directly assessed from $D$. In short, the MDL score of a pair $(X_i, U)$ is the number of bits required to encode information about $X_i$ and its parents $U$ if we were to use Huffmann coding of $D$.

To derive the bound we exploit the following observation. When $U$ is empty, we have the maximal conditional entropy $\mathcal{H}(X_i|\emptyset)=\mathcal{H}(X_i)$ and the minimal network complexity $\mathcal{NC}(X_i,\emptyset) = 0.5\cdot\log_2(m)\cdot(r_i - 1)$. By increasing the size of $U$ we can decrease conditional entropy of $X_i$, which has the theoretical limit of~$0$, at the expense of increasing network complexity. This follows from the basic properties of entropy and the definition of the network complexity term. Once the network complexity $\mathcal{NC}(X_i,U)$ is greater or equal to $s(X_i, \emptyset) = m \cdot \mathcal{H}(X_i) + \mathcal{NC}(X_i,\emptyset)$, irrespective of which variables we add to $U$, the score $s(X_i, U)$ will always increase. This is the point at which network complexity outweighs any gains from the decreasing entropy of $X_i$. Consequently, if $U$ satisfies
\[ \textrm{\it Condition~1: }\mathcal{NC}(X_i,U) \geq s(X_i,\emptyset), \]
then any superset of $U$ can be excluded from further consideration, since it will not admit new optimal or maximal parent sets for~$X_i$. The efficiency of {\it Condition 1} depends on the input data $D$. Nevertheless, it works extremely well in practice. For example, in our experiments, reported in Section~\ref{sec:results}, we found that for real-world data with $n=70$ we never considered nodes with more than nine variables.

We can further extend our pruning strategy by using the following observation~\cite{Tian2000}. The lowest entropy we can achieve for $X_i$ is $\mathcal{H}(X_i,\calX - \{X_i\})$. Now consider the score $d(X_i,U)$. Here we have that if
\[ \textrm{\it Condition~2: } d(X_i, U) \leq m\cdot\mathcal{H}(X_i,\calX - \{X_i\}) + \mathcal{NC}(X_i,U) \] holds, then no superset of $U$ can improve the score $d(X_i,U)$. This is because any superset of $U$ has higher network complexity, and hence $\displaystyle\forall_{U'\supset U} m\cdot\mathcal{H}(X_i,\calX - \{X_i\}) + \mathcal{NC}(X_i,U) \le \mathcal{H}(X_i,\calX - \{X_i\}) + \mathcal{NC}(X_i,U')$. As previously, if $U$ satisfies the condition we can exclude it from further considerations, since it will not admit new optimal or maximal parent sets. The example effect of applying our pruning conditions is shown in Figure~\ref{fig:space_pruned}.

Although both conditions achieve the same goal of pruning the search space, they differ in which information they require. To test {\it Condition~1}, we use only network complexity, which can be computed for any pair $X_i$ and $U$ independently of other $U' \subset U$, i.e. independently of predecessors of $U$ in the dynamic programming lattice. On the other hand, {\it Condition~2} provides a tighter bound but depends on the score $d(X_i, U)$, which, as we explained earlier, requires access to the maximal parent sets of $X_i$.

\subsection{Parallel Exploration}

Because of the memory and computational complexity, which remains challenging even when our pruning conditions are applied, we focus our parallel strategy on the distributed memory systems, with the Apache Spark platform~\cite{Zaharia2016} serving as an execution vehicle.

Recall that our goal is to traverse in the top-down fashion the dynamic programming lattices for all $X_i$. A node $U$ in a lattice corresponds to a computational task that evaluates $s(X_i, U)$, tests if $U$ is a maximal parent set, and checks if supersets of $U$ can admit new maximal parent sets. These tests are the source of precedence constraints between the tasks. The main idea of our parallel approach is as follows. We on-the-fly generate and ``fold'' the dynamic programming lattices for different $X_i$ into a single lattice with lower memory requirement and denser computational load. We explore the resulting lattice in parallel, initially in the BFS mode, and switch to DFS when memory becomes a bottleneck. To store and access maximal parent sets discovered in the process, we maintain a global state, which is synchronized via reduction between the layers. Finally, we reorder computations within each layer to eliminate fine-grain synchronization between the tasks, that otherwise would be necessary to effectuate our pruning conditions. Below, we explain each element of our approach.

\subsubsection{Folding Lattices} If we consider the dynamic programming lattice for variable $X_i$, then until our pruning conditions become effective we have to manage $\binom{n-1}{l}$ tasks at the level $l$ of the lattice. Consequently, the memory required to represent the entire layer $l$ is bounded by $B_1 = c_1 \cdot n \cdot \binom{n-1}{l}$, assuming cost $c_1$ to store a task. This easily becomes problematic for larger problems as soon as $l > 2$. The problem persists even when pruning takes place, since initially only some of the tasks are removed from consideration. However, we can ``fold'' the dynamic programming lattices such that the tasks sharing the same set $U$ across different lattices are represented by a single task (see Figure~\ref{fig:space_folded}). Let the memory taken by such combined task be $c_2$. The memory requirement of the new lattice is $B_2 = c_2 \cdot \binom{n}{l}$. This gives us $\displaystyle\frac{B_1}{B_2} = \frac{c_1}{c_2} \cdot (n - l)$ reduction in memory complexity. To store a task we can use a bitmap, where $i$-th bit indicates whether element $i$ is in a set. In such case, $c_2 = 2 \cdot c_1$, since in the compacted lattice we require one bitmap to represent all $X_i$ for a task, and one bitmap to represent the actual $U$ (v.s. storing only $U$ in the original lattice). By using bitmaps we additionally reduce memory overhead, and we can realize basic operations, like testing set inclusion, with only few hardware instructions. The memory reduction becomes less significant as $l$ increases. However, this is acceptable, since we expect that  thanks to the pruning conditions the search process will terminate early, which we confirm via experimental results.

The main advantage of our ``folding'' step is significantly increased computational density. To process a single task in the ``folded'' lattice, we have to perform multiple evaluations of $s$ with the same parent set $U$. Without explaining details of how $s$ is computed from $D$, we note that by having the same parents in the consecutive invocations of $s$, we can precompute statistics about $D$ induced by $U$, and reuse them from one invocation to another. Consequently, the cost of processing a task in the ``folded'' lattice is higher than the cost of processing an individual corresponding task in the original lattice, but it is lower than the total cost of processing all corresponding tasks from the original lattices, i.e. if $X$ is a set of random variables sharing $U$ we have that $T(s(X, U)) < \sum_{X_i \in X} T(s(X_i,U))$, where $T$ is the processing~cost.

By ``folding'' the lattices, we limit parallelism in the first two stages of the lattice. However, this has a negligible effect on the scalability, since even for large $n$ the cost of processing these layers is minimal compared to the total processing time. Alternatively, we may decide to ``fold'' the lattices only after the desired level of parallelism is achieved. Finally, the computational cost of individual tasks becomes non-uniform, but this is addressed by the dynamic scheduling at the run-time.

\subsubsection{Limiting Synchronization}\label{sec:sync}

Consider the task for node $U$ at the layer $l$, and suppose that {\it Condition~1} or {\it Condition~2} holds for $U$. In such case, no task that corresponds to a superset of $U$ should be generated and included in the layer $l + 1$, as it will not contribute new maximal parent sets. In other words, at given layer we should see only those tasks whose predecessors all did not satisfy the pruning conditions. However, to enforce this requirement we would need either complex synchronization between all tasks within the same layer, or a reduction operation on all possible tasks for the next layer, which effectively would defy the purpose of pruning.

To address this problem, we can change the way in which tasks for the next layer are enumerated, such that synchronization is bypassed at the small cost of considering a few unnecessary tasks in the next layer. We first order variables in $\calX$ by the decreasing number of states they have in $D$, i.e. for any $X_i$ and $X_j$, if $i < j$ then $r_i \ge r_j$, and we maintain this ordering for every node~$U$. If two variables have the same number of states, we use $\mathcal{H}(X_i, \calX - \{X_i\}) < \mathcal{H}(X_j, \calX - \{X_j\})$ as a secondary condition. Then, when deciding whether a task should be considered in the next layer, instead of checking if any of its predecessors satisfied pruning condition, which would require synchronization, we check only one selected predecessor. Specifically, let $X_j \in U$ be the maximal element in $U$. To enumerate descendants of $U$, we consider only $U' = U \cup \{X_k\}$ for all $k > j$. Thus, node $U$ becomes a predecessor to $n - j$ nodes~(see Figure~\ref{fig:space_reordered}). At the same time, from Eq.~(\ref{eq:nc}) and {\it Conditions~1}~and~{\it 2}, it follows that smaller the $j$ the higher the probability that $U$ will satisfy pruning conditions. To see why, observe that the network complexity term grows as the product of the number of states that variables in $U$ can assume. Because variables are ordered by the decreasing number of their states and the increasing entropy, we have that if $|U| = |U''|$ and $j < j''$, where $X_j$ is the maximal element in $U$ and $X_{j''}$ is the maximal element in $U''$, then $\mathcal{NC}(X_i,U) \geq \mathcal{NC}(X_i,U'')$. Consequently, nodes that are predecessors to the largest number of nodes in the next layer are most likely to meet the pruning conditions. While this approach does not guarantee that all tasks that should be pruned will not be generated, it works very well in practice. In fact, in our experiments we found that we remove no less than $97\%$ of all tasks that should be pruned. The remaining $3\%$ constitute an extra work of processing nodes that do not contribute maximal parent sets. Note that these nodes once processed never create successors and thus the extra work overhead does not propagate.

To decide whether node $U$ at layer $l$ is a maximal parent set for $X_i$, we require optimal parent set scores, $d(X_i, U')$, for all subsets $U' \subset U$ from the layer $l - 1$. As we explained earlier, instead of maintaining a complete list of all optimal parent sets, to retrieve $d(X_i, U)$ we depend only on the previously enumerated maximal parent sets. For each variable $X_i$, we store a list $L(X_i)$ of its maximal parent sets represented by tuples $(U, s(X_i, U))$, and sorted by the score $s(X_i, U)$. Then to extract all optimal parent set scores for $X_i$ and $U$ we require $O(|L(X_i)|)$ scan of $L(X_i)$. This is affordable, since even complete $L(X_i)$ is very small for a typical input data (see Table~\ref{tab:data} in Section~\ref{sec:results}). Each task at layer $l$ may contribute a new maximal parent set that must be available to all tasks for $X_i$ in the subsequent layers. Consequently, we maintain all $L(X_i)$ as a global state that is updated via all-to-all reduction, with list merging as an operator, after given layer is entirely processed. This step can be efficiently executed considering a small size of $L(X_i)$.

\subsubsection{Changing Exploration Mode}\label{sec:DFS}

While our pruning conditions significantly constrain the search space, for large problems the number of the tasks generated in the later stages of the execution may still exceed the available main memory. This in turn would lead to the undesired out-of-core execution. After processing all tasks at layer $l$, we can estimate the number of tasks that layer $l + 2$ will have in the worst case. If that number exceeds the total available memory, it is reasonable to conclude that we have sufficient parallelism, and instead of creating new tasks we can change the mode of execution into a memory preserving~DFS. Specifically, for each node $U$ at layer~$l + 1$, instead of considering all supersets of $U$ independently, we can process them sequentially following the DFS order (see Figure~\ref{fig:space_reordered}). However, in such case we cannot assume that the global list of all maximal parent sets is consistent between different tasks. As a result, some tasks could end up generating incorrect maximal parent sets or could perform extra work because without the access to the complete list of maximal parent sets {\it Condition~2} could unnecessarily fail. To mitigate this situation, we flag all maximal parent sets generated in the DFS mode that potentially could not be maximal in the global sense, i.e. when maximal parent sets from other tasks are taken into the account. These are maximal parent sets $U$ whose at least one strict subset $U' \subset U$ has been processed in a different task. Once all tasks are processed, we perform reduction to obtain the final global state for all $L(X_i)$. Then, we proceed with checking if the flagged maximal parent sets remain maximal in the merged $L(X_i)$. Let $z_f(X_i)$ be the total number of maximal parent sets flagged when running in the DFS mode. The cost of verifying these maximal parent sets is $O(z_f(X_i) \cdot |L(x_i)|)$. This is because, in the worst case, for every flagged maximal parent set $U$, we have to check if none of the remaining elements in $L(X_i)$ is a subset of $U$ with a better score. However, it turns out that in the practical settings $z_f(X_i)$ is a very small number, and in fact frequently we have that $z_f(X_i) = 0$. To understand why, consider the following. The memory requirements due to BFS grow exponentially with the depth of the dynamic programming lattice. At the same time, because the network complexity term, $\mathcal{NC}$, grows exponentially as well, the probability of enumerating new maximal parent sets decreases as we progress to the higher layers of the lattice. In our experiments, for all tested data sets we did not enumerate new maximal parent sets beyond layer $l=6$. At the same time, if the available main memory is limited, and we are forced to switch to the DFS mode early, then we can expect that the majority of the maximal parent sets tested by a single task will not be depending on the maximal parent sets discovered in other tasks. This is a direct consequence of the precedence constraints within the~lattice.

When switching to the DFS mode, we can expect an increased computational imbalance between the tasks. However, the largest tasks which could be the source of the most significant imbalance are the ones which are the most likely to be pruned. At the same time, the number of the DFS tasks will remain sufficient to provide room for load balancing at the run-time, which we confirm by experiments.

\subsection{Apache Spark Implementation}\label{sec:spark}

We implemented our parallel approach using the Apache Spark platform. The reason we use Spark is purely pragmatic: the platform supports locality-aware dynamic task scheduling, which we directly benefit from, since our computational tasks can be heterogeneous owing to the lattice ``folding'' and the potential use of the DFS mode. Additionally, Spark API makes expressing iterative BSP-style programs extremely productive. While it is clear that using one of the traditional HPC models, e.g. MPI or UPC, we could probably achieve faster implementation, we believe that the scalability would remain comparable.

The high level exploration components of our method are implemented in Python, and the computationally intensive parts, specifically evaluations of function $s$, are offloaded to the efficient, SIMD-parallel C++ kernel derived from our \textsf{SABNA} package~\cite{Karan2016,SABNA}. Apache Spark is usually regarded as a platform for the data intensive computing. In our case, the input data is typically very small (i.e. at the order of MB), however, it quickly generates massive new data representing individual tasks of the dynamic programming lattice. Below we provide details of the implementation assuming that reader is familiar with the basics of the Apache Spark interface~\cite{SPARK}.

We follow the standard BSP model realized via iterative transformations on a sequence of Spark Resilient Distributed Datasets (RDDs), where $RDD_i$ represents layer $i$ of the compacted dynamic programming lattice. To represent a node in the lattice, RDD stores a tuple $(X, U)$, where both $X$ and $U$ are expressed via bitmaps, and $X$ keeps variables that share $U$. To obtain $RDD_i$, we initialize and parallelize $RDD_0$ on Spark driver, since this is very inexpensive operation. Then, we iteratively apply the following transformations: $RDD_i = RDD_{i-1}.repartition(p).mapPartitions(M)$, where $p$ is a small multiple, usually four (as suggested in several Apache Spark best practices), of the total number of cores that Spark executors can use, and $M$ is the mapping function that: 1)~evaluates function~$s$ for all variables in $X$, and identifies potential maximal parent sets, 2)~checks the conditions to constraint the search space, and 3)~accordingly yields nodes for the next layer to explore. The repartitioning transformation is to ensure good load balancing between executors since the number of tasks grows from one RDD to the next. Here we depend on the default Spark scheduler. We use $mapPartitions$, instead of a more natural $map$, to enable indexing of the data $D$ when map $M$ is initialized. By indexing $D$ we significantly accelerate computations of $s$, and by doing so only once per partition we avoid unnecessary overheads. At the end of every iteration we materialize RDD by invoking Spark's $count$ action. Based on the resulting size of RDD, we assess the memory requirement for the subsequent iterations, and decide whether we should be switching to the DFS mode. Finally, at any point of the execution we make sure that the last two RDDs are cached and remain in the main memory to avoid expensive RDD recomputing or restoring from the secondary storage.  

The mapping function $M$ makes use of the information about maximal parent sets from previous iterations, i.e. $L(X_i)$ for all $X_i$. To maintain this global state we use a combination of Spark accumulator and broadcast variables, since the memory cost of representing maximal parent sets is very small. In a given iteration, newly discovered maximal parent sets are added by each executor to a customized Spark accumulator to form an update to the global list of all maximal parent sets. As this could lead to potential duplicate entries in $L(X_i)$ when a task fails or speculative execution is enabled, we make sure that only unique entries are considered. After the $count$ action at the end of the iteration is performed, the accumulator is reduced an the global list managed by the Spark driver is updated and broadcast back to the executors. Together with the $count$ and $repartition$ step, these operations represent communication and synchronization stages in the BSP model.

In the DFS mode, instead of generating RDD for the next layer, which would exceed the available main memory, we apply another transformation to the current RDD, where we explore each partition as described in Section~\ref{sec:DFS}. As we explained earlier, this increases the computational cost of every task and makes tasks more heterogeneous. However, at this stage the number of RDD partitions and the distribution of their computational load is such that the Spark run-time can easily maintain load balance.

\section{Experimental Results}\label{sec:results}

To understand performance characteristics of our approach, we performed a set of experiments on a standalone Apache Spark cluster with 25 nodes and GbE interconnect. Each node in the cluster is equipped with 20-core dual-socket Intel Xeon E5v3 2.3~GHz processor, 64~GB of RAM and a standard SATA hard drive. The shared file system is run under GPFS, however, this is of minor importance considering that the input data is very small, even for the largest considered problems, and it is accessed only once at the very beginning of the computations. In all tests, Spark driver runs with the default parameters, and since it is a very light-weight process, it is collocated with one of the executors. We allocate one Spark executor per node using the default configuration for the number of cores, i.e. each executor uses all available cores in a node. All executors are configured to use 58GB out of the available 64GB, with the remaining memory available to the operating system and Python interpreter. We note that we tested different configurations of executor-to-core ratios, across different data sets, without noticeable difference in performance.

We used several popular benchmark data sets from the UCI Machine Learning Repository~\cite{UCI}, including Alarm (AL), Hail Finder (HF), the US Census Data (UCSD) and HEPAR II. These are commonly considered too challenging to be solved exactly using sequential techniques, and are among the most demanding tests for the parent set assignment. The properties of all data sets are provided in Table~\ref{tab:data}, including: $r_i$ -- the number of states (arity) that the variables in the set can assume, $z$ -- the total number of generated maximal parent sets, and the properties of the output collection of the maximal parent sets $L(X_i)$. 

\begin{table}
\caption{Data sets used in experiments.}\label{tab:data}
\begin{tabularx}{\columnwidth}{lllXlX}
\toprule
Name        &   $n$ & $m$     &  $r_i$\newline min/max & $z$ & $|L(X_i)|$\newline min/max\\
\midrule
AL-4K    &   37  &  4,000    &  2/4     	&  2,654  	& 1/281\\
AL-10K   &   37  &  10,000   &  2/4     	&  5,636  	& 1/648\\
HF-10K   &   56  &  10,000   &  2/11 	&  3,941 	& 4/353\\
USCD     &   56  &  10,000   &  2/18     &  44,804 	& 3/3857\\
HEPAR II &   70  &  4,000    &   2/4    	&  1,714 	& 1/381\\
\bottomrule
\end{tabularx}
\end{table}

\subsection{Scalability Tests}

\begin{figure}
  \centering
  \includegraphics{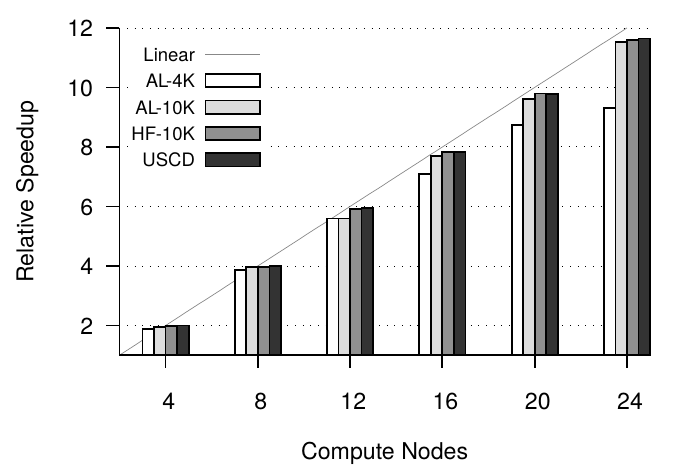}
  \caption{Relative speedup computed with respect to the execution time on two nodes.}\label{fig:speedup}
\end{figure}

In the first set of experiments, we analyzed scalability of the platform depending on the number of input variables $n$ and the number of observations $m$. We executed our Spark software on the varying number of nodes and we recorded wall time, as well as: $l_{max}$ -- the deepest processed layer in the dynamic programming lattice, $l_z$ -- the last layer at which we found new maximal parent sets, and the amount of extra work we had to perform due to removed synchronization (Section~\ref{sec:sync}). The results of this experiment are summarized in Tables~\ref{tab:time}-\ref{tab:pruning} and Figure~\ref{fig:speedup}. Here we report only relative speedup computed with respect to the time obtained on two nodes, since, except of AL-4K, we were not able to process the test data sets using sequential software.

We start the analysis by first looking at the execution time and the speedup of our method. From Table~\ref{tab:time} and Figure~\ref{fig:speedup} we can see that, with the exception of AL-4K, the software achieves very good scalability on up to 24 nodes (480 cores). In all test cases, the required main memory never exceeds 84~GB, which enables us to run completely in the BFS mode. The slightly weaker scalability for AL-4K can be attributed to the overall size of the problem. With small $n$ and relatively small number of observations the problem can be solved in a few minutes on 12 nodes. Beyond that point, the overhead of synchronization between layers becomes significant compared to the optimized compute time on the collapsed dynamic programming lattice. As the number of observations for this data set increases (data set AL-10K), computational time increases and expectedly scalability improves.

\begin{table}
\caption{Execution time in minutes.}\label{tab:time}
\begin{tabularx}{\columnwidth}{llllllll}
\toprule
& \multicolumn{7}{c}{Compute Nodes}\\
Name & 2 & 4 & 8 & 12 & 16 & 20 & 24\\
\midrule
AL-4K  & 27.1  & 14.3   & 7   & 4.8   & 4.8  & 3.1  & 2.9\\
AL-10K &   241.4    &  123.4  &   61    &  41.5   & 31.3  &  25.11   & 20.9\\
HF-10K & 425.4 & 214 & 107.5  & 71.8  & 54.2 & 43.4 & 36.6\\
USCD   & 995.7 & 496.5 & 249.2 & 167.6 & 127.1 & 101.8 & 85.4\\
\bottomrule
\end{tabularx}
\end{table}
\begin{table}
\caption{Effect of reorganizing the search space.}\label{tab:pruning}
\begin{tabularx}{\columnwidth}{XllXl}
\toprule
Name  &  $l_{max}$ & $l_z$ & Nodes processed & Extra work \\
\midrule
AL-4K  &   9     &    4    &  $4.6 \times 10^7$  &  1.68\%\\
AL-10K &   10    &    5    &  $1.6 \times 10^8$  &  1.72\%\\
HF-10K &   7     &    4   &   $3.1 \times 10^7$  &  3.2\%\\
USCD   &   8     &    6   &   $2.1 \times 10^8$  &  1.23\%\\
\bottomrule
\end{tabularx}
\end{table}

Because of the relatively small size, we were able to process AL-4K using a sequential code in 2,435 minutes. The sequential code is written entirely in C++, is optimized for memory usage and provides the same lattice constraining techniques as the parallel version. It also uses the same computational kernel to compute $s$. While this result suggests outstanding super-linear performance of our parallel code, we should keep in mind that the comparison is not completely fair, since the sequential version has significant overheads due to memory management (to avoid enumerating unnecessary tasks). Nevertheless, the result shows that even ``small'' problems can take more than a day to process sequentially, and this time can be easily reduced to minutes by using a cluster with a few nodes.

Next we consider the effectiveness of our approach in terms of removing synchronization and constraining the search space. Table~\ref{tab:pruning} shows that in the worst case we have to perform only $3.2\%$ of extra work compared to the completely synchronized version in which no unnecessary tasks are generated. At the same time, the total number of processed nodes is a small fraction of what we would have to process without constraining the search space. For example, for the USCD benchmark the total number of tasks up to layer $l_{max} = 8$ is 1,689,096,333 when no pruning is applied, and it is reduced to approximately $12\%$ of that when the pruning is enabled. Even then however, the total number of tasks to process is of the order of $10^8$, which demonstrates complexity of the problem and the need for parallel approach. The same pattern holds true for other tested data~sets.

The last layer at which we enumerate new maximal parent sets, $l_z$, is always much smaller than $l_{max}$. This suggests that there is a room to tighten the pruning conditions. At the same time, it confirms that switching to the DFS at the higher levels of the dynamic programming lattice will not trigger any significant work due to how we manage the global state with all maximal parent sets. Finally, by looking at the results for AL-4 and AL-10 data sets, we can see that both $l_{max}$ and $l_{z}$ are increasing when the number of observations increases. This is because with the growing $m$ the network complexity term increases logarithmically for any variable $X_i$, but $s(X_i,\emptyset)$ grows linearly. Consequently, the effectiveness of the pruning conditions decreases. Nevertheless, the overall performance of the method remains reasonable. 

\subsection{HEPAR II Test}

In our second experiment, we focused on the HEPAR~II test data. This benchmark comes from one of the early clinical decision support systems for multiple-disorder diagnosis of liver that involved a complex Bayesian network~\cite{Onisko2000}. As we already mentioned, the parent set assignment problem plays a critical role in the exact Bayesian networks learning, and hence directly translates into our ability to build high quality models for critical applications. This makes the benchmark interesting from the practical point of view. The benchmark is also challenging as it contains 70 variables, and all variables assume only few states, making it hard to identify variables that should be pruned.

To process HEPAR II we used all 25 nodes of our cluster. The experiment took 20 hours and 17 minutes to complete, with $l_{max} = 8$ and $l_z = 4$. To the best of our knowledge, this is the first time exact results for HEPAR II are reported. The peak memory consumption was 327~GB, and the execution involved the total of 10,770,519,474 tasks. Because the total memory available in our cluster is 1.6~TB, we again were able to process this benchmark completely in the BFS mode. However, to see how turning into the DFS mode affects the performance, we limited the available memory to 4~GB per node or 100~GB total memory. In this case, at layer $l=7$ we had to switch to the DFS mode to process the remaining 9,427,586,763 tasks. This had the minor impact on the performance, and we were able to complete the entire execution in 20 hours and 28 minutes. Here we should keep in mind that because $l_z = 4$ there were no new maximal parent sets discovered when running in the DFS mode. However, we believe that even with new parent sets discovered the performance would not be drastically changed.



\section{Related Work}\label{sec:prior}

Because of its importance, the parent assignment problem has been considered as a standalone question~\cite{Koivisto2006,Scanagatta2015} and in the relation to the structure learning of Bayesian networks~\cite{Yuan2011,Karan2016}. In~\cite{Koivisto2006}, Koivisto provides several hardness results that suggest that the parent assignment for a single variable most likely has no polynomial-time solution. This motivates our parallel approach as there is a practical need to push the size of the problems that can be solved exactly in realistic time limits. In~\cite{Yuan2011,Karan2016}, multiple authors discuss the application of maximal parent sets in exact Bayesian networks structure learning. However, in each case maximal parent sets are assumed to be given and no details of how that is achieved are provided. In this paper, we provide the actual scalable algorithm for maximal parent sets enumeration, which in fact can be combined with any Bayesian network structure learning strategy. There is a significant body of work on solving maximal parent sets enumeration while discovering Bayesian network structure~\cite{Ott2004,Koivisto2004,Singh2005}, including parallel algorithms~\cite{Nikolova2013,Tamada2011}. However, when both problems are coupled many optimizations specific to the parent sets enumeration become infeasible. As a result, these combined approaches do not scale and are limited to the instances with 30 to 40 variables, even when using thousands of cores and provably optimal MPI-based realizations~\cite{Nikolova2013,Tamada2011}. Finally, recently Scanagatta et al.~\cite{Scanagatta2015} proposed a greedy heuristic that depends on a fast approximation of the actual scoring function to constraint the number of explored parent sets. While this approach can be used to solve problems larger than what we report, it does not provide any quality guarantees. In contrast, our method is guaranteed to provide the exact solution.

\section{Conclusion}\label{sec:conclusion}

The exact parent set identification is a challenging problem with important applications in the exact structure learning of Bayesian networks. In this paper, we proposed a new scalable distributed memory approach to the problem, and we used it to efficiently process HEPAR II data set. This experiment clearly demonstrated that our method can handle even the most challenging data sets and using only limited hardware resources. This in turn opens new possibilities for exact learning of large Bayesian networks, as with some effort our method can be combined with the already existing solvers, e.g.~\cite{Karan2016}. Our approach is scalable and we believe can be generalized to other popular scoring functions including AIC and BDeu. Since the efficiency of constraining the search space for these functions is currently unclear, the ability of our solution to adopt to heavy workloads provides a significant advantage.


\section*{Acknowledgments}
Authors wish to acknowledge hardware and technical support provided by the Center for Computational Research at the University at Buffalo.

\bibliographystyle{IEEEtran}
\bibliography{references}

\end{document}